\documentclass{article} 
\usepackage{nips15submit_e,times}
\usepackage{hyperref}
\usepackage{dirtytalk}
\usepackage{graphicx}
\usepackage{url}
\usepackage{caption}
\usepackage{subcaption}
\usepackage{makecell}

\usepackage{booktabs,tabularx}

\title{Challenges of Applying Deep Reinforcement Learning in Dynamic Dispatching}
\author{
Hamed Khorasgani \\
Hitachi  Industrial AI Lab \\
\texttt{hamed.khorasgani@hal.hitachi.com} 
\And
Haiyan Wang \\
Hitachi  Industrial AI Lab \\
\texttt{haiyan.wang@hal.hitachi.com} 
\And
Chetan Gupta \\
Hitachi  Industrial AI Lab \\
\texttt{chetan.gupta@hal.hitachi.com} 
}

\nipsfinalcopy 

\begin{document}

\maketitle

\begin{abstract}
Dynamic dispatching aims to smartly allocate the right
resources to the right place at the right time. Dynamic dispatching  is  one of the core problems for
operations optimization in the mining industry. 
Theoretically, deep reinforcement learning (RL) should be a natural fit to solve this problem. However,
the industry  relies on heuristics or even
human intuitions, which are often short-sighted and
sub-optimal solutions. In this paper, we review the main challenges in using deep RL to address the dynamic dispatching problem in the mining industry.
\end{abstract}

\section{Introduction}
The mining industry is on the cusp of an unprecedented digital transformation which is  
focused on embracing  technologies such as artificial intelligence (AI) and the Internet of Things (IoT) to improve operational efficiency, productivity, and safety~\cite{lala2016prod}. 
While still in its early stages, the adoption of AI-based advanced analytics is already reshaping the industry by lowering maintenance costs, decreasing downtimes, and boosting outputs and chemical recoveries \cite{mckinsey2018}. 
However,
the potential of AI   extends far beyond that. 
In this paper, we presents the  potentials and challenges  of utilizing deep reinforcement learning (RL)  for Open-Pit Mining Operational Planning (OPMOP), an NP-hard problem \cite{souza2010hybrid} which seeks to balance the trade-offs between mine productivity and operational costs. While OPMOP encapsulates a wide range of operational planning tasks, we focus on the most critical: the dynamic allocation of truck-shovel resources~\cite{chaowasakoo2017}. 
Dynamic  allocation of tens of trucks to achieve maximum overall production is a collaborative multi-agent problem. 
Multi-agent deep RL algorithms have shown superhuman performance in  game environments  such as the Dota 2 video game \cite{berner2019dota} and StarCraft II \cite{vinyals2019alphastar}. In the mining industry, millions of dollars  can be saved by small  improvements in   productivity.  The unprecedented performance of multi-agent deep RL in learning sophisticated policies in collaborative and competitive games and the huge potential benefits in the  mining industry  makes one wonder why the industry is reluctant in using   multi-agent deep RL  algorithms to solve the  dynamic dispatching problem.  Recently, Dulac-Arnold et al.  \cite{dulac2019challenges} listed nine main challenges in applying deep RL in real life applications. 
In this work, we  discuss some of these challenges in more details for the dynamic dispatching application. Moreover, we discuss 
 two additional challenges in using deep RL to  address the dynamic dispatching problem:
 \begin{itemize}
    \item {Multi-agent system with variable number of  agents. Truck failures or truck repairs can change the number of agents during  a shift.}  
     \item {Variable goals and constraints  and the cost of retraining. Goals such as desired production level,  or constraints such as maximum speed could change in a short period of time, and it is not always feasible to retrain the model in a timely way. }  
  
   
\end{itemize}
  Table \ref{table:results} presents a summary of the presented challenges in \cite{dulac2019challenges} in the dynamic dispatching context, and the two additional challenges we introduced in this paper. 
  
   

\begin{table}
\caption{Practical Challenges of Deep RL in Dynamic Dispatching.}
\begin{tabularx}{\textwidth}{c*{3}{>{\raggedright\arraybackslash}X}}
 \textbf{Presented Challenges} \cite{dulac2019challenges}
  &              \textbf{Details} \\
\midrule
\midrule
\textit{Off line training} &  must be addressed to remove  the need for expensive simulators.\\ 
  
\midrule
\textit{Sample efficiency} 
  &  must be addressed to make off line training possible.  \\
\midrule
\textit{High-dimensionality}  & must be addressed   in  large scale multi-agent environments such as  mines.   \\ 
  
  \midrule
\textit{Safe exploration}   & must be addressed  moving toward  self-driving trucks.  \\ 
  
\midrule
\textit{\makecell{Non-stationary \\ $\&$ stochastic \\ environments} }
  &    must be addressed  because  of  inaccurate simulators. \\
\midrule
\textit{Unspecified reward}  & must be addressed   because   mines  typically have several goals and  constraints. \\
  
  \midrule
\textit{Explainability} &must be addressed so the  operators can trust and validate  the policies. 
\\ 
  
  \midrule
\textit{Real-time}  & is not  a bottleneck for  dynamic dispatching. \\ 
  
\midrule
\textit{Delays}
  &  must be addressed  especially delays in the reward feedback in a  multi-agent environment.  \\
\midrule
\midrule
 \textbf{Additional Challenges}
  &  \textbf{Details} \\
  
    \midrule
    \midrule  
 \textbf{\textit{\makecell{Variable number of  agents } }}&  must be addressed because of trucks' failures and repairs.  
\\ 
  
  \midrule
  \textbf{\textit{\makecell{Variable goals \\ $\&$ constraints}} } &  must be addressed   because quick retraining is not  feasible. \\ 
  

\bottomrule
\bottomrule
\end{tabularx}
\label{table:results}
\end{table}
   
Section \ref{Problem Formulation} formulates the dynamic dispatching  problem in the mining industry and  presents the current algorithms the industry applies to solve this  problem. 
Section \ref{Practical Challenges of Deep RL} presents the challenges the industry faces in applying deep RL algorithms. Section \ref{Conclusions} concludes the paper. 

\section{Problem Formulation and the Current Industry Practices}
\label{Problem Formulation}

\begin{figure} [t]
           \begin{subfigure}[b]{0.6\textwidth}
         \centering
         \includegraphics[width=\textwidth]{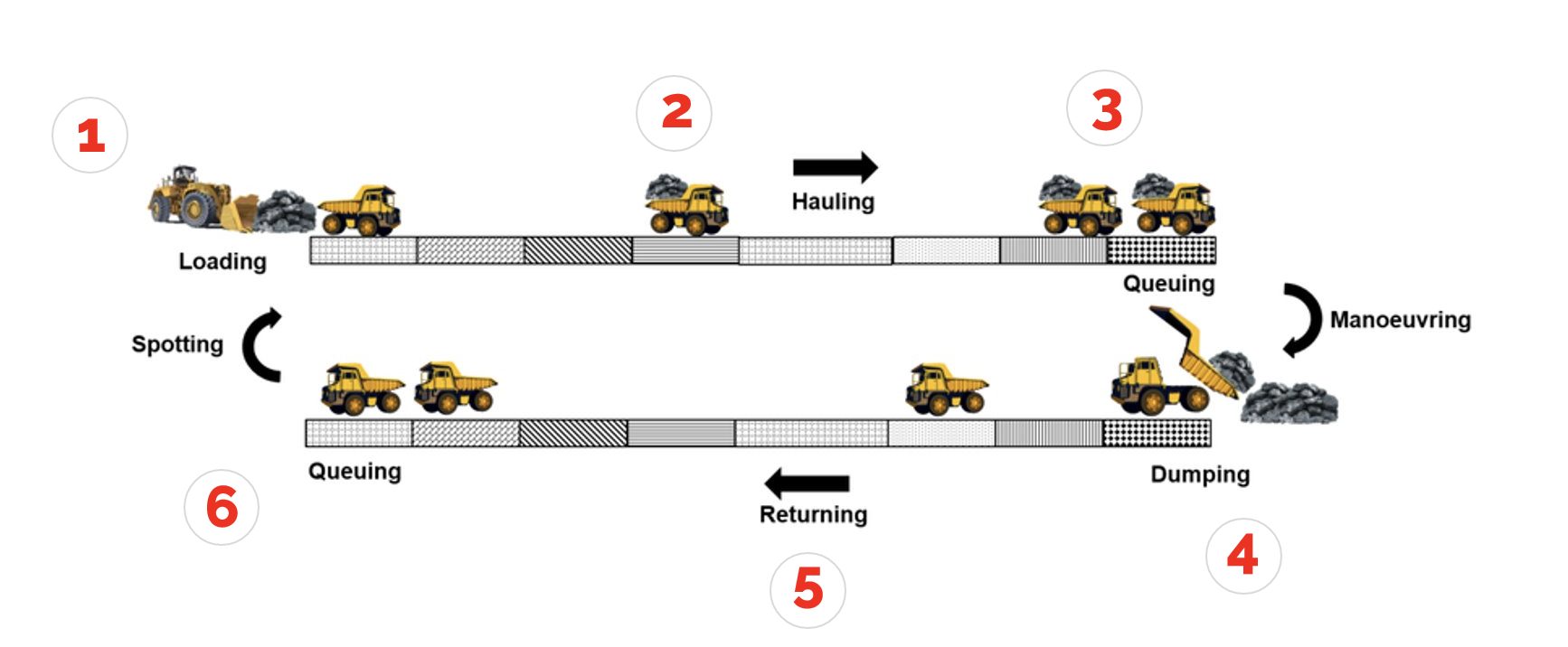} 
         \caption{}
        \label{fig:mining_cycle}
     \end{subfigure} \vspace{-5pt}
        \begin{subfigure}[b]{0.4\textwidth}
      \includegraphics[width=\textwidth]{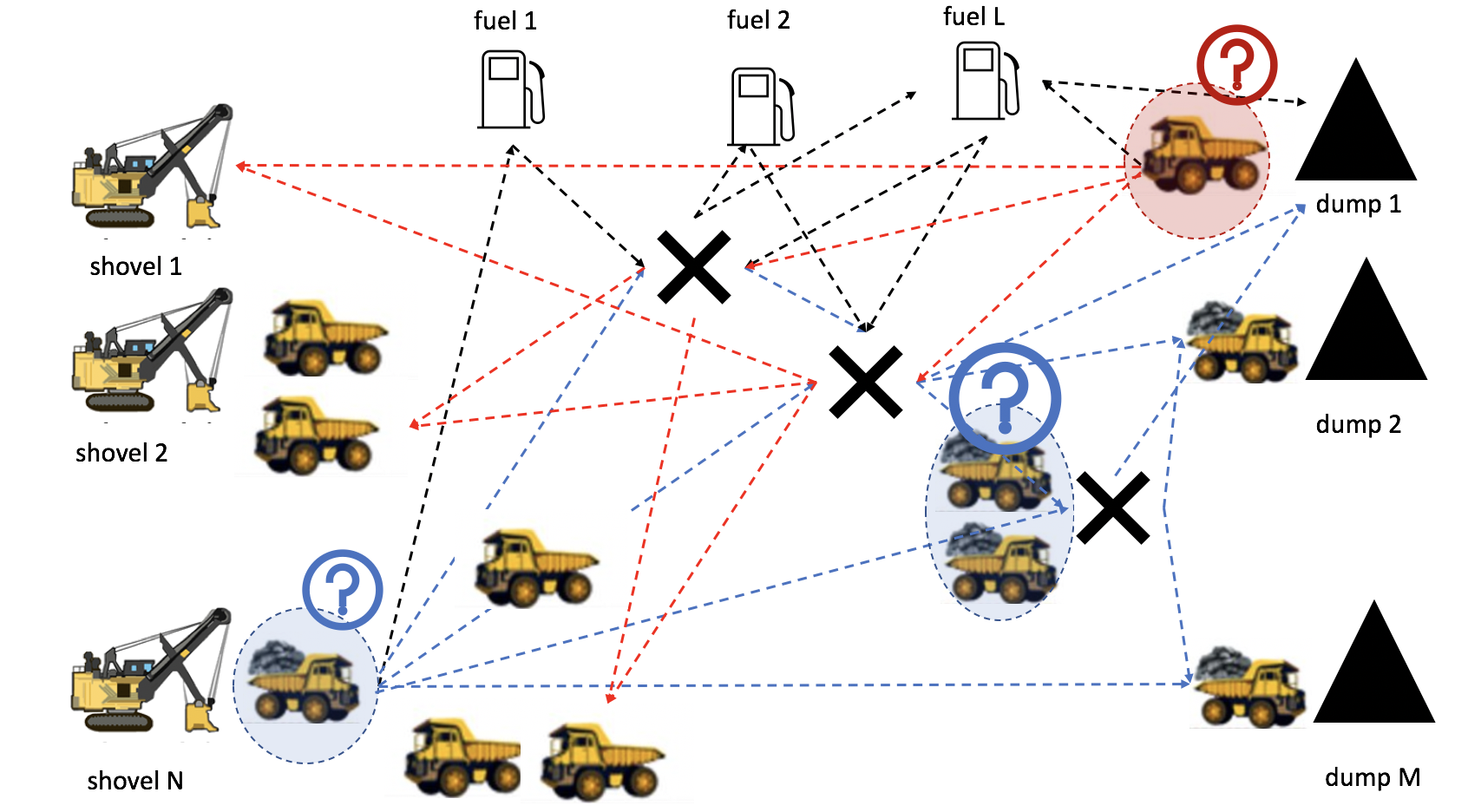}
        \caption{}
      \label{fig:dp_problem}
    \end{subfigure}
         \caption{\small{(a) Truck activities in one complete cycle in mining operations, namely loading, haulage, and maneuvering, dumping    (a dump represents a crusher,  or a waste dump), driving back empty, and spotting the shovel. (b) Graph representation of dynamic dispatching problem in mining. When trucks finish loading or dumping (highlighted in dashed circles), they need to be dispatched to a new dump, shovel or a fuel station destination.}}
        \label{fig:problem}
        \vspace{-20pt}
\end{figure}

In the open-pit mine operations, dispatch decisions orchestrate trucks to shovels for ore loading, and to dumps for ore delivery. This process, referred to as a truck cycle, is repeated continually over a 12-hour operational shift. Figure~\ref{fig:mining_cycle} illustrates the sequence of events contained within a single truck cycle. Moreover, when a truck's fuel level is low, it must make a trip to a fuel station. 
An additional \textit{queuing} step is introduced when the arrival rate of trucks to a given shovel, dump, or fuel station exceeds its loading, dumping, or fueling rate. 
Queuing can also occur at crossroads  or even in the  middle of a road because of the road condition, accidents or truck failures. 
Queuing represents a major inefficiency for trucks resulting in a drop in productivity. Another inefficiency worth noting occurs when the truck arrival rate falls below the shovel loading rate. This scenario is known as \textit{shovel starvation}, and results in idle shovels. Consequently, the goal of a good dispatch policy is to minimize both starvation for shovels and queuing for trucks. 


The industry  uses two main approaches  to allocate trucks to the shovels and dumps: 1) fixed allocation, and 2) dynamic allocation \cite{sadri2008development}. Typically, the fuel dispatching is dealt with separately and each truck with low fuel is assigned to the closest fuel station. 
In the fixed allocation strategy, each truck is allocated to a fixed shovel   for  the entire shift. In these algorithms, the main challenge is to determine the optimum   number of trucks with proper capacities  allocated to  each shovel and dump.  Fixed allocation algorithms do not require on-line computation and communication and therefore, they are more suitable for mines with less computer network infrastructure. However, these approaches are often less efficient compared to dynamic allocation algorithms where the trucks are  dispatched to the shovels and dumps based on the current conditions of the mine. After the destination point is determined, typically a   search-based algorithm is applied to find the best route between the initial point and the destination.  The search takes into account all operational and equipment constraints \cite{ahangaran2012real}.
Shortest Queue (SQ) is one of the most popular  dynamic allocation approaches \cite{subtil2011practical}. SQ aims to reduce the overall cycle times by dispatching  each truck to the destination with the minimal number of waiting trucks, including the en-route trucks. Intuitively, shorter cycle time  leads to more cycles and, therefore,  higher production. However, this may not be true when we have heterogeneous trucks with different capacities.  Smart Shortest Queue (SSQ) or Shortest Processing Time First (SPTF) take advantage of the activity time predictions to estimate the waiting time for the current truck to be served and allocates the truck to the destination with minimum  waiting time. The SSQ's  goal is  to minimize the actual serving time, which is often difficult to achieve for conventional SQ since it does not have the activity time estimation capability \cite{rose2001shortest}.


 Existing heuristic rules such as Shortest Queue (SQ)~ and Shortest Processing Time First (SPTF) rely on short-term and local indicators (e.g., waiting time) to make decisions, leading to short-sighted and sub-optimal solutions. It is possible to formulate the dynamic dispatching problem as a multi-agent RL problem as follows.
 \begin{itemize}
     \item {Agents: any dispatchable truck can be considered as an agent. Truck fleets can be composed of trucks with varying haulage capacities, driving speeds, loading/unloading time, etc., resulting in truck fleets with heterogeneous agents.}  
     \item {Environment: shovels, dumps, fuel stations, roads and crossroads  are the environment for each agent. The shovels and dumps can have different capacities meaning the number of trucks that each can serve simultaneously is different. The environment  changes gradually as the mine progresses. }  
    \item {States: in addition to the agent's  state such as location, and fuel level, each truck typically has access to the global state of the mine at each moment. 
The  global state  includes the location of each truck and its relevant attributes such as capacity and speed.  Moreover, each agent typically has access to the estimation of the global state in near future.}
   \item {Action space: the action space for this problem encapsulates all possible actions available to each agent. Since the dispatch problem inherently tries to determine the best shovel/dump to send a truck, each unique shovel and dump within the mine represents a possible action.
Moreover, when a truck's fuel is low, the dispatching algorithm has to determine the most efficient  fuel station for refilling. Finally, at the crossroads, each truck has to decide to go forward or wait for other trucks to pass first.  Based on this approach, the action space for each truck is  a finite and discrete space.}

  \item {Goal: the overall goal is to avoid starvation, and long queues to maximize the mine production. In addition to the main goal, the mine typically has additional goals and constraints  such as visiting each dump site frequently enough to prevent dry ore. Moreover, the movement of dust control trucks, which  are  large water trucks to spray haul roads, may limit the hauling  trucks' movements and further complicate the dynamic dispatching problem.}

  
   
\end{itemize}


\section{Practical Challenges of Deep RL}
\label{Practical Challenges of Deep RL}
In this section, we first discuss well-known challenges of the application of RL in real life presented by Dulac-Arnold et al.  \cite{dulac2019challenges} from a mining  perspective. Next, we present two additional challenges  that we believe the industry should overcome to make RL a viable solution.

\subsection{Well-known challenges}
Unlike game environments  \cite{berner2019dota,vinyals2019alphastar}  or even some real world applications such as the robotic arm \cite{finn2015learning,pinto2016supersizing}, it is not possible to train RL algorithms in the real world when it comes to the mining. Safety has been discussed in the literature as the main obstacle in training deep RL in real world applications. In   applications such as autonomous  robots, the hazard caused from exploration during the training is the main issue preventing us from learning deep RL policies  in the real world and several solutions have been proposed for safe exploration for RL \cite{gu2017deep,dalal2018safe}. 
In dynamic dispatching, safety is not a major obstacle in real world training. 
Despite efforts to automate the mines, the trucks are still driven by human operators and therefore,   there is no significant safety issue  associated with  allocating a truck to a wrong shovel or dump for exploration. 
However, the pure cost of such an experiment makes it impractical. Each mine typically operates in 8-hour or 12-hour shifts and the goal of dynamic dispatching is to maximize the overall production during each shift. Recently, there have been significant improvements in sample efficiency of deep RL. Techniques such as \textit{replay buffer} value estimation methods \cite{mnih2015human,lillicrap2015continuous},  \textit{imagination rollouts} \cite{gu2016continuous}, and  \textit{model-based guided policy search} \cite{deisenroth2011pilco} 
 have improved  sample efficiency of deep RL algorithms significantly.
However, even the most efficient deep RL requires hundreds or thousands of episodes before converging, and in dynamic dispatching, that would be equal to asking the mine management to have the full mine operating in suboptimal capacity for days or even years just to come up with an efficient policy. No matter how great the potential gain can be, this is simply too high of a cost  for an operating mine to pay.

Typically, the mining industry has access to plenty of off-line operation data. The most advanced  mines are  connected to the cloud and their operation data history  is recorded with a  high frequency. These datasets include the location of each truck and their loads  at every few seconds. In fact, most dispatching  algorithms use the historical data to estimate different variables such as  waiting time for each shovel or dump, or traveling time for each truck, given its load and the road condition\cite{ristovski2017}. Since the mines are already using their historical data to learn and update different supervised deep learning models  in a periodic manner, one may ask why the industry is not using the off-line data to learn optimal policies using deep RL.  Fujimoto et al.  \cite{fujimoto2019off} showed  that the extrapolation
error  prevents standard deep RL algorithms 
such as deep Qnetwork (DQN)  \cite{mnih2015human} and Deep Deterministic Policy Gradients (DDPG) \cite{lillicrap2015continuous} to learn efficient policy from off-line datasets.  To avoid the extrapolation
error, they proposed  to select the optimal actions among  the actions which  are likely from  the off-line dataset. 
 In a similar analysis, Kumar et al.   \cite{kumar2019stabilizing} identified a bootstrapping error as the source of instability in off-line deep RL. 
They defined the bootstrapping error as the error in value estimation generated due to bootstrapping from actions  outside of the training data distribution.  To solve  the bootstrapping error, Kumar et al.   \cite{kumar2019stabilizing}  proposed an RL algorithm which ensures that the learned policy matches the   distribution of  the off-line dataset actions. Yu et al.  \cite{yu2020mopo} proposed an off-line model-based deep RL which learns an ensemble of system's dynamic models  and assigns  negative reward proportional to  the error between the estimated models. Higher error between the learned models shows that the agent is entering  unseen environments and the negative reward helps the agent learn the optimal policy while staying  in the off-line data distribution. Even though some progress has been made in off-line deep RL, the technology is still in its early stages and to the best of our knowledge there has been no work that has applied off-line RL successfully to a large-scale multi-agent environment such as dynamic dispatching. The large action space of a mine with hundreds of trucks makes   the success of an off-line RL for these systems much less likely. 

With high cost of training in a real mine, and limitations of off-line learning, using simulators to learn deep learning policy is the only practical approach. Developing and maintaining  accurate simulators is expensive and sometimes infeasible. 
Real mines are often much more complicated  compared to the developed simulators. Changes in the weather and road conditions, complications associated with operating humans in the loop, and possible truck failures make what agents may experience in the real mine much more  complex. 
There have been several attempts to address the uncertainties in the simulators. These approaches typically consider MDP  formulation, model noise and uncertainty  as bounded unknown variables and optimize the network for the worst case scenario \cite{mankowitz2018learning,shashua2017deep}. 
The obvious problem with these methods is that by considering
the worst case scenario, 
 the solutions tend to be very conservative.  
 Derman et al.  \cite{derman2018soft}  proposed an algorithm which focuses on the uncertainty distributions instead of the worst-case scenario to avoid the overly conservative policies. Peng et al.  \cite{peng2018sim} used dynamics randomization to randomizes a robotic arm simulator 
parameters such as mass, friction, and time-step between actions in each episode in a simulator during the training  to learn a robust policy, and successfully applied the policy to the real robotic arm. Even though their results seem promising, it is much harder to generate  realistic perturbations in a mining simulator with hundreds of human operators in the loop. 
In  dynamic dispatching,  we may not be able to even model uncertainties using additional parameters in many cases. For example, it is common that we encounter a traffic  jam in a real mine because of weather,  road conditions, or an accident. A traffic jam can   change the entire dynamic model of the mine and cannot be modeled with simple parameters. A possible solution could be to combine real world data with incomplete simulators to generate robust solutions.

\subsection{Multi-agent system with variable number of  agents}
The dynamic dispatching problem is a multi-agent problem. 
When the number of agents is small, it is possible  to model multi-agent problems  using  a centralized approach 
where we 
train a centralized policy  over  the agents'  joint
observations  and output a joint set of actions. One can imagine that this approach does not scale well for dynamic dispatching  and  we will quickly have state and action space with very large dimensions.  A more realistic approach 
 is to use an autonomous learner for each agent such as independent DQN~\cite{mnih2015human} which distinguishes agents by identities. 
Even though the independent learners  address the scalability problem   to some extent, 
they  suffer from  convergence point of view as the environments become  non-stationarity. In fact, these algorithms  model the other agents as part of the environment and, therefore,  the policy networks have to chase  moving targets as the  agents' behaviors  change during the training  \cite{tsitsiklis1994}.
To address the convergence problem, centralized
learning with decentralized execution approaches  have been proposed in recent years. In these methods, a  centralized learning approach is combined with a 
 decentralized execution mechanism to have the best of both worlds.  Lowe et al.  \cite{lowe2017multi} proposed multi-agent deep deterministic policy gradient (MADDPG), which includes   a centralized critic network and  decentralized actor networks  for the agents.  Sunehag et al.  \cite{sunehag2018value} proposed a linear additive  value decomposition approach where  the total Q value is modeled as a sum of 
 individual agents' Q values.  Rashid et al.  \cite{rashid2018qmix} proposed  Q-MIX network, which allows  a richer mixing of Q agents compared to the linear additive   value-decomposition.  

Even though the centralized
learning with decentralized execution approaches have shown promising results in many applications, they are not practical solutions to address the dynamic dispatching problem for the mining industry. In the dynamic dispatching problem, the 
number of trucks are not constant. The number of available trucks can change at each given day and even when the number of trucks is known ahead of time, it is fairly common for a truck to break during the operation and becomes unavailable for the rest of the  operating shift.    Solutions such as Q-MIX  \cite{rashid2018qmix}  and value decomposition \cite{sunehag2018value} assume that the number of agents are fixed. Removing an agent leaves a hole in the network. Moreover, we cannot add additional agents during an episode, which can be the case in the mining industry. 
A trivial solution to this problem is to learn different policy networks for different number of agents, so we can inact the appropriate policy when the number of agents changes.  However, 
the range of agents could  be very large in a mine. Moreover, in addition to the truck failures, we may have failures in  the shovels and crushers (dumps)   as well. To have a model trained and updated  for each combination  is expensive or even infeasible   especially for a large mine with hundreds of trucks, and tens of shovels and crushers.

Foerster et al.  \cite{foerster2016learning}  proposed a  single network with shared parameters to  reduce the number of learned parameters and speed up the learning. Having a shared policy among agents solves the agent failure challenge, however it can make the problem non-stationarity. To address the non-stationarity problem  that
can occur when multiple agents learn concurrently, they  disabled experience replay. 
 Disabling the experience replay can weaken the sample efficiency and  slow down the learning process. 
Wang et al.  \cite{wang2018actor} introduced a new embedding  state representation  in  actor-critic framework to address  the variable quantity of agents. 
Because truck failures are inevitable in the mining industry, the necessity to deliver optimum dispatching with fewer trucks is far more critical than  delivering optimum dispatching with more trucks.  Traditionally,  fault tolerance has been a key research area in multi-agent systems (MAS). In theory, when an agent is failed, other agents with similar capabilities
can reorganize to compensate for the loss \cite{mellouli2007reorganization}.
One of the missing parts in research and  validation of multi-agent RL algorithms is fault tolerance qualities of the solutions. The industry is much more likely to  adopt deep RL policies if they present a reliable   fault tolerance   strategy (optimum solution with fewer trucks).

\subsection{Variable goals and constraints, and the cost of retraining}
Mining Operations are a very dynamic environment with plans constantly needing to be updated to reflect changes in the budget, production targets, mill requirements, maintenance planning as well as all the unplanned events.  
The industry needs to be able to make  changes within a  real-time environment so that the results being generated reflect the current status of the operation. Deep RL algorithms typically require many samples during  training (sample complexity). Sample complexity leads to high computational time and costs. Computational cost and time can be justified for the mine as a one-time charge. However,   
  oftentimes small changes in the system goal,  such as  changing the  desired production target, or a new constraint  such as  limited truck's speed due to road condition,   require retraining the model. Moreover, 
  mines often have several short-term and long-term objectives. For example, the goal is not only to maximize the production but also to move enough material to each dump to avoid having idle crushers.

Meta-RL and multi-tasks RL algorithms which aim to learn  multiple  tasks and  the ability to adapt efficiently to new  tasks have become  very popular in recent years and can potentially address some of  these challenges.  Fin et al.  \cite{finn2017model} proposed to learn a shared general policy across tasks during the training. During the test time, their algorithm can adapt to new tasks with few observations. Rakelly et al.  \cite{rakelly2019efficient}  proposed to learn a latent representation of the tasks during the training. The  representation learning  makes it possible to solve new tasks in the test time.
Lin et al.  \cite{lin2020model} proposed a model-based adversarial meta-RL algorithm which optimizes the policy for the worst-case sub-optimality (the hardest task). By using the adversarial network,   the algorithm learns acceptable solutions even for the hardest tasks. 
Even though significant progress has made in recent years, the researchers have been focusing on narrow and simple tasks  mostly in robotic domain \cite{yu2020meta}. 
The mining industry needs a multi-task RL algorithm that can learn a variety of tasks with possibly different distributions. For example, it should be able to deliver vastly different production levels which possibly could lead to different number of required trucks (agents). Moreover, we need to address the constraints in an efficient manner. For example, we may have to set a maximum  trucks' speed in specific roads due to the roads' conditions or the operation of dust control trucks in those roads and the policy still has to  deliver an optimum dispatching solution without retraining.

\section{Conclusions}
\label{Conclusions}
In this work, we presented main challenges in applying deep RL to the dynamic dispatching problem in the mining industry. We discussed 1) the infeasibility of training on real mines even though, safe exploration is not a main challenge in the mining industry, 2) the challenges of off-line training for the mining industry, such as large state and action space and low chance of convergence, and 3) the challenges raised from incomplete simulators which lead to non-stationary and stochastic environments. We skipped some common challenges such as  1) unspecified reward functions,  2)  lack of explainability, 3) real-time requirements and 4) delays, as they were similar in the mining industry and other industries. Finally, we introduced two new challenges that have to be addressed for the industry to embrace deep RL algorithms: 1) variable number of  agents in deep multi-agent RL and 2) variable goals and constraints. We believe this paper helps to bridge  the gap between the scientific researches and the industrial demands. 

\section*{Acknowledgments}
We thank Tengyu Ma, Ahmed Farahat and Hsiu-Khuern Tang for their  very valuable  suggestions and comments on the paper. 
\bibliographystyle{ieeetr}
\bibliography{bibliography}

\clearpage

\end{document}